\documentclass{article}
\usepackage{spconf,amsmath,graphicx,mathtools,breqn,array,caption}

\usepackage{xcolor}
\title{Volumetric landmark detection with a multi-scale shift equivariant neural network}
 \name{Tianyu Ma$^{\star}$ \qquad Ajay Gupta$^{\dagger}$ \qquad Mert R. Sabuncu$^{\star\dagger}$\thanks{This research was funded by NIH grants 1R21AG050122, R01LM012719, R01AG053949; and, NSF CAREER 1748377, and NSF NeuroNex Grant1707312.}}
 \address{$^{\star}$ 
            School of Electrical and Computer Engineering; and Meinig School of Biomedical Engineering\\
            Cornell University \\
     $^{\dagger}$
        Department of Radiology \\
        Weill Cornell Medical College}
\begin{document}
%
\maketitle
\begin{abstract}
Deep neural networks yield promising results in a wide range of computer vision applications, including landmark detection.
A major challenge for accurate anatomical landmark detection in volumetric images such as clinical CT scans is that large-scale data often constrain the capacity of the employed neural network architecture due to GPU memory limitations, which in turn can limit the precision of the output. 
We propose a multi-scale, end-to-end deep learning method that achieves fast and memory-efficient landmark detection in 3D images. 
Our architecture consists of blocks of shift-equivariant networks, each of which performs landmark detection at a different spatial scale. 
These blocks are connected from coarse to fine-scale, with differentiable resampling layers, so that all levels can be trained together. 
We also present a noise injection strategy that increases the robustness of the model and allows us to quantify uncertainty at test time.
We evaluate our method for carotid artery bifurcations detection on 263 CT volumes and achieve a better than state-of-the-art accuracy with mean Euclidean distance error of 2.81mm.                   
\end{abstract}
\section{Introduction}
\label{sec:intro}

Recent advances in deep convolutional neural networks (CNNs) show great potential in various medical image computing applications, such as segmentation~\cite{segnet}, registration~\cite{dalca2019unsupervised}, and landmark detection~\cite{regressing}. 
In this study, we consider the problem of identifying landmark points (sometimes called key-points) that correspond to certain anatomical and/or salient features in the image.
This problem has been explored with traditional machine learning techniques~\cite{dantone2012real,criminisi2010regression,uvrivcavr2012detector}, yet more recently deep learning methods have produced the best results~\cite{ranjan2017hyperface,zhang2016joint}. 
One common approach with deep learning is to use a full-convolutional architecture such as U-Net~\cite{unet} to compute a heatmap image at the output that highlights the location of the landmark(s).
Thus landmark localization is turned into an image-to-heatmap regression problem~\cite{regressing}, where the ground truth coordinates are used to place Gaussian blobs (of often arbitrary size) to create training data.

An important challenge in applying deep neural networks to volumetric data is computational, since resources needed to train on large-scale volumes will be proportional to the size of data~\cite{volumetric}.
This is often addressed by breaking the problem down to multiple steps, each solved separately.
For example, one can tile the volume into 3D patches and apply landmark detection in each patch separately.
Path-level predictions will be limited by the field of view and aggregation of patch-level results into the whole volume can be challenging.
One complementary multi-stage strategy is to first use a coarse-scale network that produces candidate patches based on a lower resolution volume~\cite{volumetric,attention,deep}.
In these approaches, each step's model is trained independently, which can compromise final accuracy due to error accumulation.

In this paper, we build on the multi-scale approach, yet propose a strategy to implement it in a single step.
Our building block is a shift-equivariant network that combines a U-Net and a center-of-mass layer to produce predicted landmark coordinates in a volumetric image of a computationally manageable size.
Blocks operating at different scales are then connected through differentiable resampling layers, which allows us to train the whole network end-to-end.
Unlike heatmap based approaches, the proposed model directly computes spatial coordinates for the landmarks. 
Our multi-scale network is memory efficient. The high resolution image which cannot fit into our GPU memory in the single-scale implementation can be used as the input for our multi-scale approach
Contrary to other methods that use a coarse-to-fine multi-scale design, we employ a single step training scheme which improves accuracy.
We further implement noise injection between scales to increase the robustness of our model and achieve uncertainty in predictions at test time.

We apply our method to computed tomography angiography (CTA) scans of the neck to localize the bifurcation of the left and right carotid arteries. 
In our experiments, we compare the proposed method to several baseline methods, where our model yields superior accuracy with useful uncertainty measurements that correlate with localization error.

\section{Proposed Method}
\label{sec:method}
\begin{figure}[htb]

\begin{minipage}[b]{1.0\linewidth}
  \centering
  \centerline{\includegraphics[width=\linewidth]{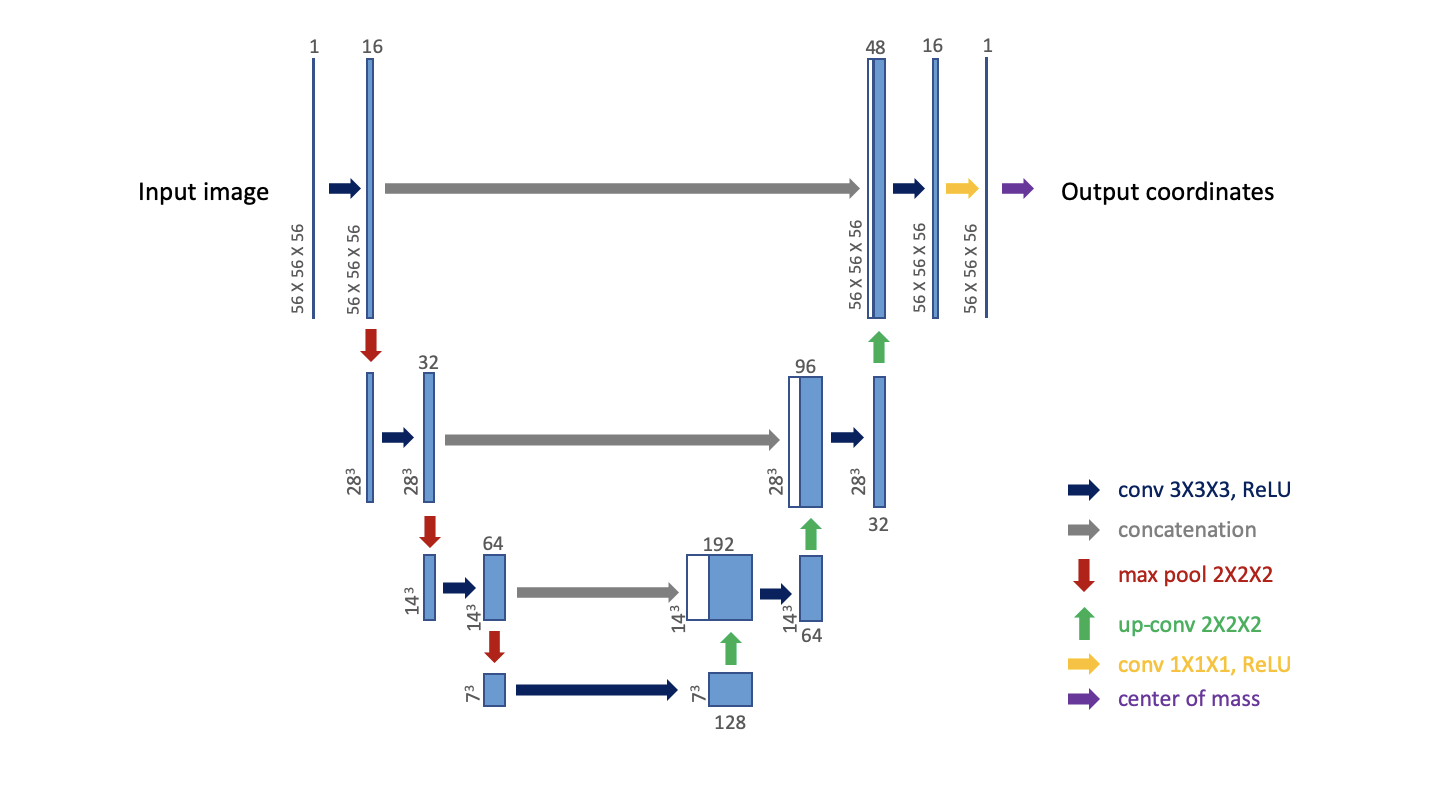}}
\end{minipage}
\caption{Localizer Network (Loc-Net): our building block.}
\label{fig:locnet}
\end{figure}

Our model for landmark detection consists of sequentially connected blocks of localizer networks (Loc-Net) that operate at different spatial scales - from coarse to fine.
Each Loc-Net is simply a 3D U-Net-like architecture~\cite{cciccek20163d} that has a center of mass (CoM) layer at the end (see Fig.~\ref{fig:locnet}).
As was demonstrated recently~\cite{sofka2017fully}, a CoM layer computes the weighted average of voxel grid coordinates, where the weights are the pixel values of the U-Net output, which we can interpret as a heatmap that highlights the landmark.
The weighted average coordinate vector is the predicted location of the landmark, computed by the scale-specific Loc-Net.
Since the U-Net is a fully convolutional architecture, and the CoM is shift equivariant, it is easy to show that the Loc-Net will be shift equivariant~\footnote{This is approximately true, since down-sampling and pooling operations compromise the shift equivariance. A recent paper proposes a way to fix this issue~\cite{zhang2019making}.}.
That is, if we shift the input image by a certain amount (along the three axes), the predicted landmark coordinates will be shifted by the same amount.
This, we believe, should be a critical property of a landmark detection model.

Another advantage of the CoM layer is that it circumvents the need for treating the problem as heatmap regression - a common approach for detecting landmarks with neural networks~\cite{regressing}.
The main drawback of a heatmap regression approach is that the training annotations, which are often provided as coordinates, need to be converted into heatmaps, e.g., by placing Gaussian kernels.
These kernels are usually of arbitrary size. 
If they are too small, training the networks can be hard. 
If they are too big, the precision of the landmark detection can be compromised.

\begin{figure}
\begin{minipage}[b]{1.0\linewidth}
  \centering
  \centerline{\includegraphics[width=\linewidth]{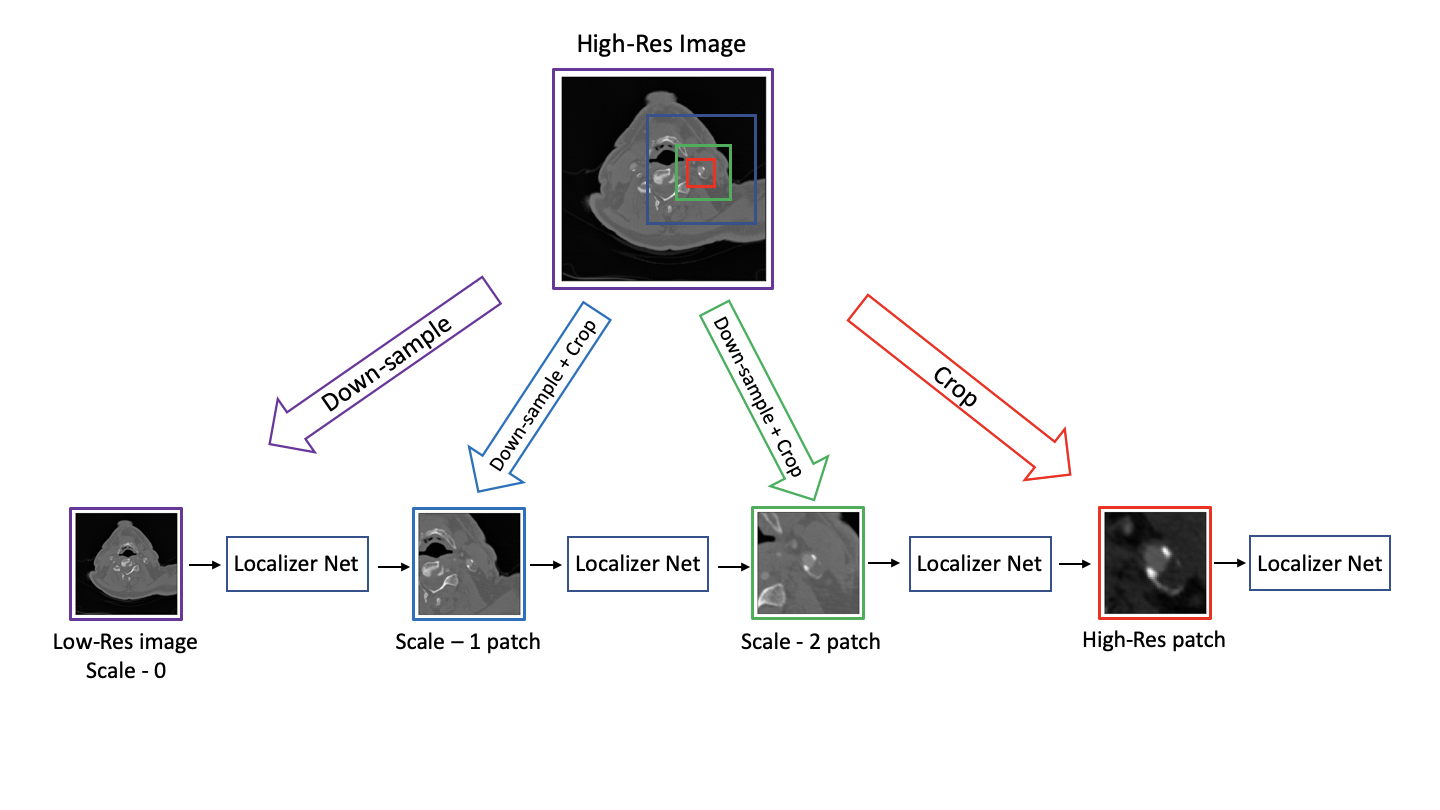}}
\end{minipage}
\caption{Proposed Multi-scale Landmark Localization. \textcolor{black}{The architecture for all Localizer Nets is shown in Fig.~\ref{fig:locnet}}}
\label{fig:all}
\end{figure}

We connect scale-specific Loc-Net modules sequentially, from low to high resolution, via an intermediate differentiable crop and resample layer. 
This layer creates a 3D patch of size $56 \times 56 \times 56$, centered around the predicted landmark location and sampled at the next spatial scale.
Fig.~\ref{fig:all} illustrates our framework.
This model is then trained end-to-end via minimizing the loss between the predicted location of the landmark and the ground truth.
Our loss is a weighted sum of the scale-specific squared loss functions. 
Note that each scale-specific Loc-Net computes a prediction of the landmark location, which we can evaluate with a squared loss with respect to the ground truth.
In our experiments, we implemented a piecewise linearly scheduled weighting scheme in combining scale-specific losses.
All weights of the scale-specific loss terms were constant during each epoch.
The relative weight of the lowest-scale loss started at one and linearly decreased to zero by epoch 500.
The relative weight of the highest-scale loss started at zero and linearly increased to 1 by epoch 500.
The loss weights corresponding to the other two scales started at zero, linearly increased and peaked at epoch 250, and then linearly decreased to zero by epoch 500.
Thus, initially, our model's training was focused on predicting the landmark coordinates at lower scales, eventually focusing on the accuracy of the highest resolution.


In this paper, we also explore the use of a novel stochastic regularization technique, which allows us to improve test time accuracy and quantify uncertainty in predictions.
Instead of propagating the predicted landmark to the final scale, we add random shifts (see below for further details).
This approach is reminiscent of stochastic dropout~\cite{srivastava2014dropout} and its variants. 
While dropout was originally proposed to be used during training (to improve robustness), it was recently shown that its use at test time can be viewed as an approximate Bayesian inference strategy that can yield well-calibrated uncertainty estimates - allowing researchers to combat the over-confidence issue of modern deep neural networks~\cite{gal2016dropout}.
Inspired by this, we also implement our noise injection strategy at test time to compute the uncertainty in the landmark predictions.
So at test time, we implement multiple forward passes through our network with different noise realizations injected at the input of the final Loc-Net module, thus computing multiple predictions. The standard deviations of these predictions can be used to quantify the uncertainty in the localization.  


\subsection{Implementation}
We implement our code in tensorflow.
Our mini-batch size is $1$ 3D CTA volume, and all methods are trained for 500 epochs using the Adam optimizer with a learning rate of $0.0005$. For noise injection, a $5$mm uniform random shift in each direction is applied before cropping the $0.5$mm resolution image patch.
\section{Empirical Results}
\label{sec:exp}

In this study approved by the Weill Cornell institutional review board (IRB), we analyze 263 anonymized CTA exams obtained in the course of clinical care from patients with unilateral $>50\%$ extracranial carotid artery stenosis, who were imaged at Weill Cornell Medicine/New York Presbyterian Hospital.
\textcolor{black}{All the data involving human subjects used in this paper has obtained the corresponding ethics approval/waiver.}
All CTA volumes are interpolated to a high resolution grid of size $448 \times 448 \times 448$ and $0.5 \textrm{mm}$ voxel spacing. All down-sampled images used by our model are then created from this set of high resolution images. The lower resolution images have $1 \textrm{mm}$, $2 \textrm{mm}$ and $4$mm isotropic voxels. The data are randomly split into three non-overlapping sets where 167 for training, 42 for validation and 54 for testing. All of our landmarks for left and right carotid artery bifurcations are manually annotated by a trained human (T.M.). 

To demonstrate the efficacy of the CoM layer, as a baseline, we implement a single scale 3D U-Net with a similar architecture to our Loc-Net~\footnote{We included an additional convolution plus down-sample layer} that computes a heatmap output.
This is trained with L2-loss on the heatmap images, where the ground truth heatmaps were created by centering a Gaussian kernel (with std = 6 mm) at the annotated landmark coordinates
This is compared to a single-scale Loc-Net, which has the CoM layer at the end.
Due to limited GPU memory, these single-scale models accept $1$mm spacing $224 \times 224 \times 224$ sized volumes as input.

We consider other baselines as well.
Our state-of-the-art (SOTA) baseline is a patch based technique that was recently proposed~\cite{patch}. We also implement two alternative versions of our proposed multi-scale model. 
In the first alternative, we disconnect the Loc-Nets from the different  scales and train them sequentially in separate steps.
In this training scheme, the first scale Loc-Net, for example, receives no learning signal from the highest resolution.
A second alternative is the full multi-scale model trained in an end-to-end fashion, but without noise injection. 

\begin{figure}[htb]

\begin{minipage}[b]{1.0\linewidth}
  \centering
  \centerline{\includegraphics[width=8.5cm]{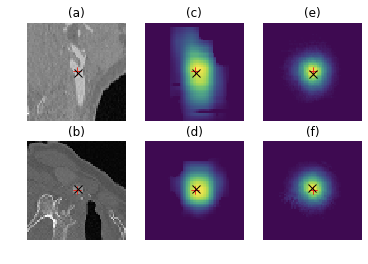}}
\end{minipage}
\caption{A representative test case. '{\color{black}+}' shows the ground truth, and 'x' indicates the prediction of landmark. (a) and (b) are  sagittal and axial views; (c) and (d) are the heatmaps obtained with single-scale Loc-Net with CoM layer; (e) and (f) are heatmaps computed with single-scale U-Net model.}
\label{Figure 1.}
\end{figure}

Table \ref{Table 1.} shows the average Euclidean distance error ($\pm$ standard deviation) for all considered methods, computed on the test data. 
First, we observe that the center of mass layer we use in our model boosts the performance of the U-Net over a heatmamp regression method. 
Fig. \ref{Figure 1.} illustrates representative examples of heatmaps computed with and without the CoM. 
In addition to a more accurate prediction, the CoM layer yields heatmaps that correlate better with the underlying anatomy.

\begin{table}
\centering
\resizebox{\columnwidth}{!}{
\begin{tabular}{ ||c|c || } 
\hline
Architectures & Error \\ 
\hline\hline
U-Net (Heatmap) ($1$mm resolution) & $6.45 \pm 9.18$mm  \\ 
\hline
Single-scale Loc-Net ($1$mm resolution)& $5.52 \pm 5.69$mm \\ 
\hline
Patch-based SOTA Baseline~\cite{patch} & $4.35 \pm 4.33$mm \\
\hline
Multi-scale Loc-Net (multi-step) & $3.72 \pm 4.20$mm \\ 
\hline
Multi-scale Loc-Net (end-to-end) & $3.61 \pm 3.17$mm \\ 
\hline
Multi-scale Loc-Net (end-to-end + noise)  & $2.81 \pm 2.37$mm \\ 
\hline
\end{tabular}
}
\caption{Experimental results for different models}
\label{Table 1.}
\end{table}

The results further illustrate that the proposed multi-scale model can outperform the state-of-the-art method~\cite{patch}.
Furthermore, both end-to-end training and noise injection improve the accuracy of the predictions.

\begin{figure}[htb]

\begin{minipage}[b]{1.0\linewidth}
  \centering
  \centerline{\includegraphics[width=8.5cm]{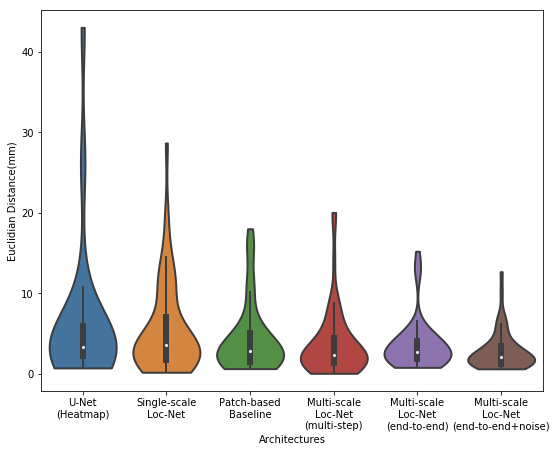}}
\end{minipage}
\caption{Test error distribution for different methods}
\label{Figure 2.}
\end{figure}

Fig. \ref{Figure 2.} presents the violin plots for all the different methods we test, which allows us the appreciate the full distribution of errors on the test cases.
We note again that the proposed method of end-to-end training the multi-scale model with noise injection yields the most compact distribution of errors with the smallest mean and median values.
The improvement in the average error is mainly due to the reduction of the failure cases. 


\begin{figure}[t!]

\begin{minipage}[b]{1.0\linewidth}
  \centering
  \centerline{\includegraphics[width=8.5cm]{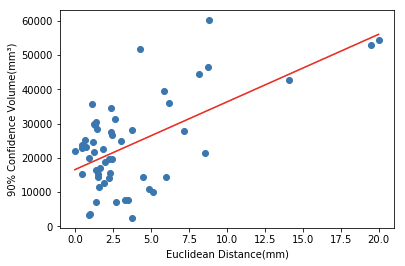}}
\end{minipage}
\caption{Quantifying localization uncertainty. We use random shifts at test time to obtain 90\% confidence volumes for our predictions (see text for further details). These volumes are correlated with the localization errors with respect to the ground truth annotations.}
\label{Figure 3.}
\end{figure}

Similar to dropout~\cite{gal2016dropout}, we  apply random shifts (noise  injection) at test time to obtain multiple predictions via 50 forward passes through the model. 
We then compute the standard deviations of these predictions along the three axes.
Note that a single forward pass through our model takes 615 milliseconds on a NVIDIA Titan Xp GPU.
Assuming an uncorrelated multivariate Gaussian, we compute the $90\%$ confidence volumes for the predictions computed by our model.
Fig. \ref{Figure 3.} plots these values with respect to the localization errors on the test cases. 
We observe that the confidence volumes (or rather the localization uncertainty) are significantly correlated (Pearson's correlation 0.608, P-value 1.08e-06) with localization error, suggesting that one could use such an approach for automatic quality assurance and propagating uncertainty to downstream analyses.


\section{Conclusion}
\label{sec:conclusion}

In this paper, we explore the use of a novel multi-scale shift-invariant model, trained in an end-to-end fashion, for automatically localizing landmarks, particularly in large-scale 3D images. 
Our model is memory efficient and thus can be deployed on very large clinical scans at full-resolution, without compromising precision.
In our experiments, we demonstrate that the employed center of mass layer can boost performance over a heatmap regression method, while simultaneously yielding more interpretable heatmaps. 
We also show that the end-to-end training of the connected multi-scale Loc-Nets can yield a significant increase in accuracy.
Our method also compares favorably to a state-of-the-art patch-based landmark localization method.
Finally, we demonstrate the use of noise injection at test time to obtain useful confidence/uncertainty estimates.

An important weakness of the proposed method is that we assume that the target landmark exists in every single query volume. This is the case for the application we consider, yet it might not hold in general. We plan to pursue this direction in the future. 




\bibliographystyle{IEEEbib}
\bibliography{regressing}

\end{document}